# Quantifier Elimination for Statistical Problems

**Dan Geiger and Christopher Meek**
Microsoft Research
Redmond 98052-6399, WA
dang@cs.technion.ac.il *, meek@microsoft.com

## Abstract

Recent improvements on Tarski's procedure for quantifier elimination in the first order theory of real numbers makes it feasible to solve small instances of the following problems completely automatically: 1. listing all equality and inequality constraints implied by a graphical model with hidden variables. 2. Comparing graphical models with hidden variables (i.e., model equivalence, inclusion, and overlap). 3. Answering questions about the identification of a model or portion of a model, and about bounds on quantities derived from a model. 4. Determining whether an independence assertion is implied from a given set of independence assertions. We discuss the foundations of quantifier elimination and demonstrate its application to these problems.

## 1 Introduction

A variety of problems from statistics, many related to graphical models, can be cast as questions in Tarski's first order theory of real numbers and answered using procedures for quantifier elimination. We discuss the foundation of quantifier elimination and demonstrate its application to some statistical problems. Other interesting statistical problems for which Tarski's theory is applicable are described in (Fagin, Halpern, and Megiddo, 1990).

Tarski (1951) has posed and solved the following problem. Given a sentence in the first order theory of real fields where variables represent real numbers, constants are 0 and 1, function symbols represent the standard binary operations of a field $\{+, -, *\}$, and relation symbols represent the standard binary relations $\{=, \neq, <, \leq, >, \geq\}$, determine whether a sentence is true, and provide a sentence without quantifiers which is equivalent to the original sentence. For example, the sentence $(\forall x)x^2 \geq 0$ is true and the formula $(\forall x)ax^2 + bx + c > 0$ is equivalent to the formula $b^2 - 4ac < 0 \wedge a > 0$. Tarski's remarkable theorem is that such an equivalent quantifier-free formula always exists and that there exists an algorithm to construct it. This result opened the way to many applications including geometrical theorem proving and geometric modeling (Caviness and Johnson, 1998).

Tarski's algorithm, developed in the thirties, is mainly of academic interest because its complexity is non-elementary (towers of exponentials in the number of variables). Collins (1975) provided an alternative procedure for quantifier elimination over the real numbers based on, what he called, Cylinder Algebraic Decompositions (*CAD*). Collin's procedure is a major breakthrough because of its clear geometrical interpretation and its complexity (doubly exponential in the number of variables). Further improvements by Collins and Hong (1991), who developed partial CADs, and an implementation called QEPCAD, made the theory available for testing in numerous application areas. We briefly review the foundations of quantifier elimination in Section 2 and add more details in the appendix.

Graphical models provide interesting statistical problems for which no solution has been known, however, as we observe, quantifier elimination provides a solution, at least in principle, to many of these problems. We consider the following classes of problems. 1. listing all equality and inequality constraints implied by a graphical model with hidden variables. 2. Comparing graphical models with hidden variables (i.e., model equivalence, inclusion, and overlap). 3. Answering questions about the identification of a model or portion of a model, and about bounds on quantities derived from a model. 4. Determining whether an independence assertion is implied from a given set of independence assertions. We demonstrate the application of quantifier elimination to these problems in Section 3.

*Author's primary affiliation: Computer Science Department, Technion, Haifa 32000, Israel.

Another type of problems for which quantifier elimination is applicable are questions about the topology of graphical models. Recently, Geiger, Heckerman, King, and Meek (1998, 1999) introduced and analyzed a class of exponential families which they termed *stratified exponential families*. They classified graphical models as linear, curved, and stratified exponential families, according to whether the parameter space over the observables is a linear manifold, a smooth manifold, or a stratified set, and demonstrated that many graphical models with hidden variables are stratified exponential families. Cylinder algebraic decomposition can be used to construct a stratification of an exponential family. Benedetti and Risler (1990) provide such a procedure, however, it has a non-elementary complexity, which makes it impractical even for very simple examples. In Section 4 we discuss other topological properties of graphical models that one may hope to compute using CADs.

## 2 Background on Real Geometry and Tarski sentences

We consider a first order language $L_R$ consisting of constants, variables, and finite-array operation and relation symbols. The constants are 0 and 1. Variables in $L$ denote real numbers. Operations are the usual $\{+, -, *\}$ operations of a field. Constants, variables, and their composition by these operations, yield *terms*. Relations consist of the binary relations $\{=, \neq, <, >, \leq, \geq\}$ which compare two terms. *Atomic formulae* consist of $True$, $False$, $t \neq t'$, $t = t'$, $t < t'$, $t > t'$, $t \leq t'$, and $t \geq t'$ where $t$ and $t'$ are terms. The standard logical connectives $\{\wedge, \vee, \neg, \rightarrow, \leftrightarrow\}$ and quantifiers $\{\exists, \forall\}$ are used to create *formulae* from atomic formulae. A variable $x$ is *bound* in a formula $\psi$ if it is in the scope of a quantifier $\exists x$ or $\forall x$; Otherwise, it is *free* in $\psi$. A formula in a first order language that contains no free variables is called a *sentence*. In the language $L_R$, a sentence is also called a *Tarski sentence*. Two Tarski sentences $\psi$ and $\psi'$ are *equivalent* if $\psi \leftrightarrow \psi'$ is true. A formula containing no quantifiers is *quantifier-free*. A formula $\psi$ is true if the sentence obtained by a universal quantification of all free variables in $\psi$ is true.

The language $L_R$ can be used to describe many geometrical, topological, and analytical properties of polynomials over the reals. Section 3 discusses several examples that relate to graphical models. Here we employ examples related to quadratic polynomials. The formula $\exists x(ax^2 + bx + c = 0)$ says that every quadratic polynomial has a real root. This formula is false. The formula $\exists x(ax^2 + bx + c = 0) \leftrightarrow (b^2 - 4ac \geq 0)$ is true. In other words, the formula $\exists x(ax^2 + bx + c = 0)$ and the quantifier-free formula $b^2 - 4ac \geq 0$ are equivalent.

**Theorem 1 (Tarski)** *Every Tarski sentence is equivalent to a quantifier-free formula. Furthermore, given a Tarski sentence $\psi$, there exists an algorithm that constructs a quantifier-free formula equivalent to $\psi$ and determines whether $\psi$ is true or not.*

This is a remarkable theorem which highlights an important property of the real numbers. The original proof of Tarski contained an algorithm of non-elementary complexity. Collins (1975) presented an improved algorithm, based on, what he termed, *cylinder algebraic decomposition (CAD).* Collins' algorithm is rooted in a famous geometrical consequence of Theorem 1, called the Tarski-Seidenberg Theorem (e.g., Mishra, 1993, Section 8.6.3, or, Benedetti and Risler, 1990), which we now present.

Define a *semi-algebraic set* $\Sigma_\psi$ (Also called a Tarski set) to be a subset of $R^n$ of the form $\{(x_1, \ldots, x_n) | \psi(x_1, \ldots, x_n) = True\}$ where $\psi$ is a formula containing exactly $n$ free variables. The formula $\psi$ is called the *defining formula* of $\Sigma_\psi$. Equivalently, due to Tarski's theorem, a subset $V$ of $R^n$ is a *semi-algebraic set* if $V = \cup_{i=1}^{s} \cap_{j=1}^{r_i} \{x \in R^n | P_{i,j}(x) \Leftrightarrow_{ij} 0\}$ were $P_{ij}$ are polynomials in $Q[x_1, \ldots, x_n]$ and $\Leftrightarrow_{ij}$ is one of the comparison relations.

A map $f : X \rightarrow Y$ where $X \subseteq R^n$ and $Y \subseteq R^m$ are semi-algebraic sets, is called *semi-algebraic* if the graph of $f$, denoted $G(f)$, is a semi-algebraic set of $R^{n+m}$. Note that if $f$ is a polynomial map then $f$ is a semi-algebraic map because its graph can be described by $m$ polynomial equalities: $y_j - f_j(x) = 0$, where $1 \leq j \leq m$.

**Theorem 2 (Tarski-Seidenberg)** *Let $f : X \rightarrow Y$ be a semi-algebraic map. Then the image $f(X) \subseteq Y$ is a semi-algebraic set.*

The proof of this theorem only requires to observe that the image $f(X)$ is given by the following Tarski set $\{y \in Y | \ \exists x((x, y) \in G(f))\}$ where $(x, y) \in G(f)$ can be described by a Tarski sentence because $G(f)$ is assumed to be a semi-algebraic set. An immediate consequence of Tarski-Seidenberg's theorem, by choosing $f$ to be a projection map, is that a projection of a semi-algebraic set is a semi-algebraic set—an observation which is the basis of Collins' algorithm for quantifier elimination. See the appendix for details.

## 3 Sample of applications

In this section we reduce several statistical problems to the problem of validating the truth of a Tarski sentence or to the problem of finding a quantifier free formula equivalent to a given one. For each problem, we describe its history and show how the *CAD* algorithm and quantifier elimination advances the state of the art in the pursuit of its solution. We consider membership problems for conditional independence (Section 3.1), determining model equivalence, inclusion, and overlap (section 3.2), listing all equality and inequality constraints implied by a graphical model with hidden variables (Section 3.3), and model identifiability (Section 3.4).

### 3.1 The membership problem

Consider a set of variables $U = \{u_1, \ldots, u_n\}$, each with a set of possible values $D(u_i)$, a set of probability distributions $\mathcal{P}$ each with a sample space $D(u_1) \times \cdots \times D(u_n)$, a set of conditional independence statements $\Sigma$ over subsets of $U$, and another conditional independence statement $I$ also over subsets of $U$. We say that $\Sigma \models_{\mathcal{P}} I$ is true if and only if for all distributions in $\mathcal{P}$ for which $\Sigma$ holds, it is the case that $I$ also holds. The membership problem is to determine, given $\Sigma, I$ and $\mathcal{P}$, whether or not $\Sigma \models_{\mathcal{P}} I$ is true.

Some effort has been devoted to solving the membership problem, when $u_i$ are assumed to be discrete variables, and $D(u_i)$ are finite sets whose cardinality is not specified (Pearl, 1988) . Polynomial algorithms for the membership problem of marginal independence statements and of saturated independence statements have been obtained (Geiger, Paz, and Pearl, 1992; Geiger and Pearl, 1993) . The membership problems for arbitrary sets of conditional independence sets $\Sigma$, under the assumption that $D(u_i)$ are finite sets whose cardinality is not known, may not be decidable. The work by Herrmann (1995) and Studeny (1992) can be regarded as steps towards establishing this claim.

We consider three cases that can be addressed using quantifier elimination. In the first case, we assume each variable is associated with a finite set of possible values $D(u_i)$ and that the cardinality of this domain is known. Under this assumption, independence facts can be expressed as a finite number polynomial constraints. In the second case, we assume the domain is the real line $R$ and that $\mathcal{P}$ is the class of non-singular multivariate normal distributions. In this case conditional independence are expressed by asserting that the determinants of some minors of the covariance matrix are zero (e.g., Lauritzen, 1989), which again can be expressed as polynomial constraints in the parameters of the family. A third case is the combination of discrete and continuous variables assuming a conditional Gaussian distribution (e.g., Lauritzen, 1989).

For example, the following instance of the membership problem, $\{X_1 \perp X_2,\ X_1 \perp X_3|X_2\} \models_{\mathcal{P}} (X_1 \perp X_3)$, where $X_1 \perp X_3|X_2$ stands for $X_1$ and $X_3$ are independent given $X_2$, and where $\mathcal{P}$ is the class of tri-variate normal distributions over $\{X_1, X_2, X_3\}$, with a positive definite covariance matrix $(\rho_{ij})$, can be written as

$$((\rho_{ij}) \text{ is a positive definite covariance matrix}) \wedge$$
$$\rho_{12} = 0 \wedge \rho_{33}\rho_{12} - \rho_{13}\rho_{23} = 0 \rightarrow \rho_{13} = 0$$

This is a short hand notation for a formula which evaluates to true using Collins' decision procedure.

A stronger version of the membership problem, which asks whether a disjunction of conditional independence statements is implied from a given set of $\Sigma$ of conditional independence statement, has been considered in (Geiger and Pearl, 1993; Meek, 1995). A well known instance of this problem is the assertion that $\{X_1 \perp X_2,\ X_1 \perp X_2|X_3\} \models_{\mathcal{P}} (X_1 \perp X_3) \vee (X_2 \perp X_3)$ is true for multivariate normal distributions. This assertion can be written as a Tarski sentence and be validated using Collins' decision procedure.

### 3.2 Model comparison

A *model* is a set of distributions $\mathcal{P} = \{P_\gamma | \gamma \in \Gamma\}$. We assume each probability distribution in $\mathcal{P}$ is indexed with exactly one value of $\gamma$ and that $\gamma$ is an algebraic number. A *polynomial (parametric) model* is a model where $\Gamma$ is given as the image of a semi-algebraic set $\Theta$ under a polynomial map $g$, i.e., $\Gamma = g(\Theta)$. For example, a discrete Bayesian network with or without hidden variables, defines a polynomial model, where $g : R^m \rightarrow R^n$ maps the Bayesian network parameters to the joint space parameters of the distribution over the observable variables. All graphical models discussed in (Geiger and Meek, 1998) are polynomial models.

When comparing alternative models, perhaps for the purpose of understanding what features of models are distinguishable, one can compare the sets of distributions that are parameterized by the model. Several authors (Verma and Pearl, 1990; Spirtes et al., 1993) have considered the problem of model comparison on the basis of the representational strength for models that have no hidden variables. For instance, Verma and Pearl (1990) have provided a simple graphical characterization of equivalence of models. Spirtes et al. (1993) have also considered model comparison among models with hidden variables where comparisons are made only on the basis of independence facts true for the observable variables.

No method for comparing graphical models with hidden variables appears in the literature. In this section we demonstrate that various comparisons between polynomial models can be posed as Tarski sentences and solved using a decision procedure for real algebra.

A polynomial model $g$ *represents* a distribution $P$ over the observed variables if there is a value of the model parameters $\theta \in \Theta$ such that $\gamma = g(\theta)$ is the index for $P$. The set-theoretic relationship $\subseteq$ is used to define the notions of model inclusion and equality. We say that $g_1 \subseteq g_2$ if and only if

$$(\forall \theta_1)(\exists \theta_2)[\theta_1 \in \Theta \rightarrow [\theta_2 \in \Theta_2 \wedge g_2(\theta_2) = g_1(\theta_1)]]$$

where the expression $\theta \in \Theta$ is a shorthand for the equation describing the semi-algebraic set $\Theta$. Equality can be written as $g_1 \subseteq g_2 \wedge g_2 \subseteq g_1$. Clearly, a sentence for model inequality can be obtained by the addition of the negation symbol to the formula above. These relations capture the notions of model equivalence and model inclusion but not model overlap. Two models $g_1, g_2$ *overlap* if and only if

$$(\exists\theta_1)(\exists\theta_2)[\theta_1 \in \Theta_1 \wedge \theta_2 \in \Theta_2 \wedge g_1(\theta_1) = g_2(\theta_2)].$$

Since we can define inclusion, equality, and overlap with Tarski sentences we can use the decision procedure to answer these questions.

Consider, for example, the naive Gaussian graphical model with three continuous observed variables $X_1, X_2, X_3$ and one hidden variable $H$. Each of the observed variables are independent given the value of $H$; graphically this model has an edge between $H$ and $X_i$ for each $i$. The model parameters are four error terms $\epsilon_i, \epsilon_H$, four conditional means $\mu_i, \mu_H$ and three weights $\beta_i$ that describe the linear relationship between $H$ and the respective $X_i$. The edge weights $\beta_i$ and the conditional means $\mu_i, \mu_H$ take on arbitrary real values and the error terms take on positive real values. Thus, the set of model parameters is a semi-algebraic set. These parameters define a joint Gaussian distribution over the hidden and observed variables as follows

$$p(H = h, X_i = x_i) = N(h|\mu_h, \epsilon_H)\cdot \prod_{i=1}^{3} N(x_i|\beta_i h + \mu_i, \epsilon_i + \beta_i^2 \epsilon_H)$$

where $N(x|\mu,\epsilon)$ is a normal density with mean $\mu$ and variance $\epsilon$, $h$ is a value of $H$ and $x_i$ is a value of $X_i$. This model has been studied extensively in statistics (e.g., Martin and McDonald 1975, Rindskopf 1984). We call it the Heywood model.

For simplicity, we focus on the covariance structure of the Heywood model assuming that the conditional means are all zero. For this model the covariance matrix over the observed variables is as follows;

$$\begin{pmatrix} \epsilon_H\beta_1^2 + \epsilon_1 & \epsilon_H\beta_1\beta_2 & \epsilon_H\beta_1\beta_3 \\ \epsilon_H\beta_1\beta_2 & \epsilon_H\beta_2^2 + \epsilon_2 & \epsilon_H\beta_1\beta_2 \\ \epsilon_H\beta_1\beta_3 & \epsilon_H\beta_2\beta_3 & \epsilon_H\beta_2^2 + \epsilon_2 \end{pmatrix}$$

We can now compare the Heywood model to a complete Gaussian Bayesian network over three variables (i.e., a model in which no edge is missing). Writing a Tarski sentence for the assertion that the Heywood model is included in the complete model evaluates to true, and a sentence asserting the equality of the two models evaluates to false. In the next section we discuss the constraints over the covariance matrix of the Heywood model which makes this model more restrictive than the complete model. In Section 3.4, we study the identifiability of the Heywood model.

### 3.3 Constraints over observable variables

Geiger and Meek (1998) use a method called implicitation, based on Groebner bases, by which one can obtain equality constraints over the parameters of the distributions over the observable variables implied by a graphical model with hidden variables. Unfortunately, the implicitation procedure is not guaranteed to find all equality constraints, nor can it be used to explicate inequality constraints over the parameters. The main reason for these limitations is that implicitation has been developed for polynomial equations over the complex field which is not an ordered field.

In this section we illustrate the use of cylinder algebraic decompositions (CAD) to identify both equality and inequality constraints implied by a polynomial model. The answer given by the CAD procedure actually provides a proof, via Theorem 5 (see appendix), that these are the only constraints implied by the model.

The general formula for this problem is $(\exists\theta)(g(\theta) - \gamma = 0)$ where $g$ is a polynomial map between $\Theta$ and $\Gamma$. We seek an equivalent quantifier-free formula. A formula that involves $\gamma$, which can be measured, and does not mention $\theta$, which cannot always be measured. Our task is an example of the problem of algebraic curve implicitation (Caviness and Johnson, 1998).

We consider the Heywood model given in the previous section. This model has been studied in statistics because of the estimation problems that arise when the inequality constraints implied by the model are not satisfied by the observed data (e.g., Martin and McDonald 1975, Rindskopf 1984). For simplicity of presentation, we assume all error terms $\epsilon$ are known. Consequently, the formula for this example is

$$(\exists\beta_1)(\exists\beta_2)(\exists\beta_3) \quad (r_{12} = \beta_1\beta_2 \wedge r_{13} = \beta_1\beta_3 \wedge r_{23} = \beta_2\beta_3)$$

Quantifier elimination yields, using QEPCAD, the following quantifier-free formula:

$$(r_{12} < 0 \wedge r_{13} < 0 \wedge r_{23} > 0) \vee (r_{12} < 0 \wedge r_{13} > 0 \wedge r_{23} < 0) \vee$$
$$(r_{12} > 0 \wedge r_{13} < 0 \wedge r_{23} < 0) \vee (r_{12} > 0 \wedge r_{13} > 0 \wedge r_{23} > 0) \vee$$
$$(r_{12} = 0 \wedge r_{23} = 0) \vee (r_{12} = 0 \wedge r_{13} = 0) \vee (r_{13} = 0 \wedge r_{23} = 0)$$

These cells consist of the volumes where the product $r_{12}r_{13}r_{23}$ is positive, the lines where two $r_{ij}$ are zeros, and the origin (Figure 1). These are all the constraints on $r_{ij}$ implied by this model.

### 3.4 Identifiability and bounding problems

One of the reasons that polynomial parametric models are often used is because one is interested in what inferences one can make about model parameters or functions of model parameters. Typically, one assumes that the distribution over the observed variables is known to belong to a set of distributions and asks if the quantities of interest (1) are uniquely determined (*identified*) (2) take on values only in some restricted region (*bounded*).

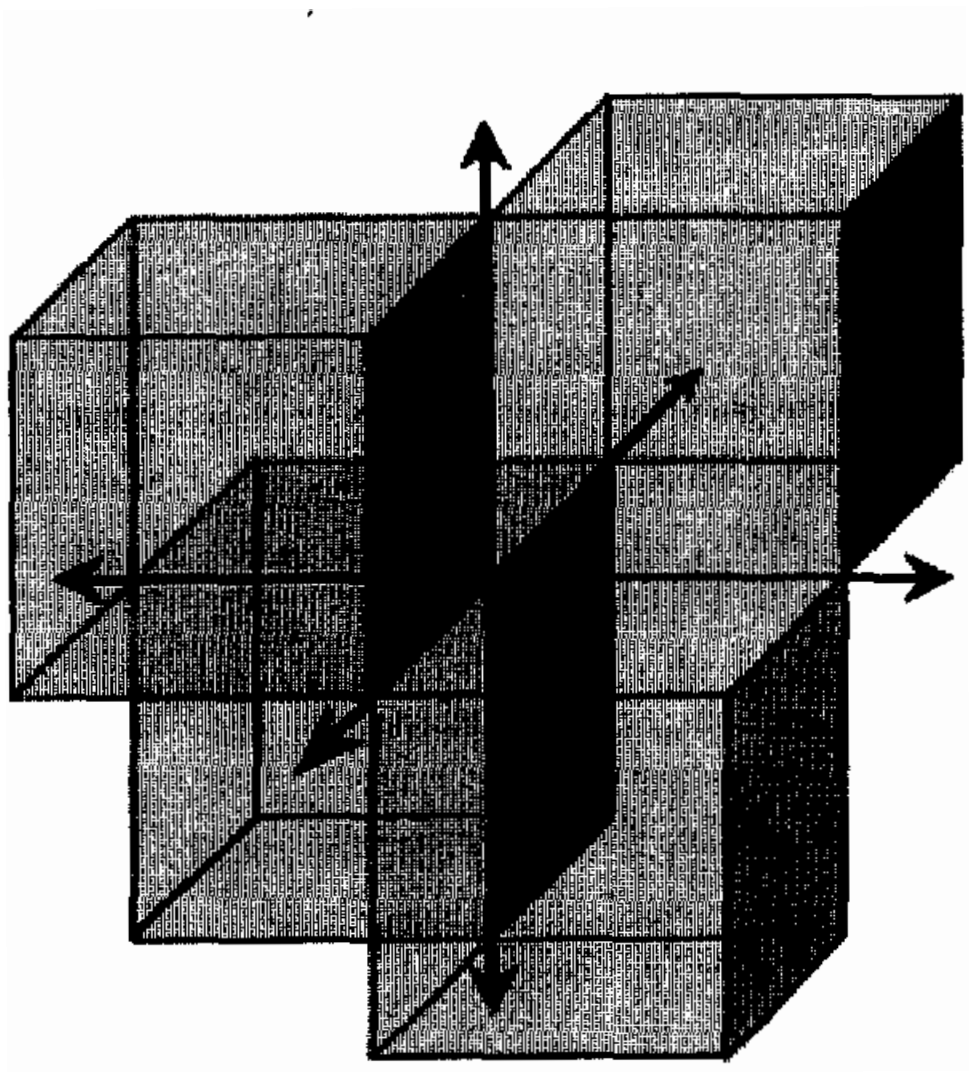

Figure 1: The *CAD* for the Heywood model.

In this section we show that one can, if the model of interest is a polynomial map $g$ with model parameters $\Theta$ and the quantity of interest $Q(\theta)$ is a polynomial function of the model parameters, use a decision procedure for real algebra and quantifier elimination to answer these questions.

First consider the question of whether $Q(\theta)$ can be identified when $g$ represents the true distribution. This question can be written as

$$(\forall\gamma)(\forall\theta)(\forall\theta')[[\theta \in \Theta \wedge \theta' \in \Theta \wedge \gamma = g(\theta) \\ \wedge g(\theta) = g(\theta')] \rightarrow Q(\theta) = Q(\theta')]$$

When $Q$ is the identity map, this question is referred to as the *global identification* problem. For the case that the true distribution $P_\gamma$ is known we eliminate the quantifier for $\gamma$ and consider $\gamma$ as a term describing the algebraic index for the distribution.

Conditions for global identifiability have been studied extensively for various types of models and in various areas of statistics (e.g., Yakowitz and Spragins, 1968; Bollen, 1989). A variety of easy-to-check sufficient conditions as well as a variety of necessary conditions for determining if a model is globally identifiable exist in the literature. Interesting examples of identification problems for causal quantities can be found in Pearl (1995).

Consider the correlational structure of the Heywood model with $\epsilon_H = 1$. The global identification question for this model can be written as follows.

$$(\forall\beta_1)(\forall\beta_2)(\forall\beta_3)(\forall\beta_1')(\forall\beta_2')(\forall\beta_3') \\ [\beta_1\beta_2 = \beta_1'\beta_2' \wedge \beta_1\beta_3 = \beta_1'\beta_3' \wedge \beta_2\beta_3 = \beta_2'\beta_3'] \\ \rightarrow [\beta_1 = \beta_1' \wedge \beta_2 = \beta_2' \wedge \beta_3 = \beta_3'].$$

This sentence is false. The parameters of this model are not identifiable whenever there is a zero covariance between two or more observed variables. Furthermore, even if all edge weights are not zero, the model is not identifiable, due to a two-fold covering of the image; if $\beta$ is a vector of edge weights, then $-\beta$ leads to the same distribution over the observables.

If one is interested in obtaining bounds for quantities when the model $g$ represents the true distribution, one can apply quantifier elimination to the formula $(\forall\gamma)(\forall\theta)\,[\gamma = g(\theta) \wedge r = Q(\theta)]$. Note that $r$ is a free variable. The result is a quantifier-free formula that describes the set of possible values for the quantity of interest, through equalities and inequalities on $r$. If the distribution is known, then we eliminate the quantifier for $\gamma$ and consider $\gamma$ as a term describing the algebraic index for the distribution. Balke and Pearl (1994) obtain bounds for a particular graphical model with hidden variables. In principle, one can obtain identical bounds automatically using quantifier elimination.

## 4 Computational topology problems

A Bayesian approach to model selection is to compute the probability that the data is generated by a given model via integration over all possible parameter values with which the model is compatible and to select a model that maximizes this probability. An asymptotic method for computing this probability (called the marginal likelihood) is the Bayesian Information Criteria (BIC) which was proven, under some regularity conditions, to be asymptotically valid for linear and curved exponential models (Schwarz, 1978, Haughton, 1988). However, for general polynomial models, $g(\Theta)$ is a stratified set, consisting of a finite collection of smooth manifolds of different dimensions, not just a single smooth manifold. For polynomial models, the asymptotic validity of BIC remains an open problem.

A ***stratified set*** is a set that has a stratification. A *stratification* of a subset $E$ of $R^n$ is a finite partition $\{A_i\}$ of $E$ such that (1) each $A_i$ (called a *stratum* of $E$) is a $d_i$-dimensional smooth manifold in $R^n$ and (2) if $A_j \cap \overline{A_i} \neq \emptyset$, then $A_j \subseteq \overline{A_i}$ and $d_j < d_i$ (frontier condition) where $\overline{A_i}$ is the closure of $A_i$ in $R^n$. See Akbulut and King (1992) for a more general definition. When $\Theta$ is a semi-algebraic set, and $g$ is a polynomial map, then $g(\Theta)$ is a stratified set (e.g., Benedeti and Risler, 1990).

We now pose several natural questions. First, can we automatically construct a stratification of $g(\Theta)$ for a polynomial model. Second, and seemingly easier, can we decide automatically whether $g(\Theta)$ is a smooth manifold. Third, can we describe automatically the set of singular points of $g(\Theta)$—points where BIC may not be valid. In the context of graphical models, it suffices to recall that a graphical model defines a polynomial model via a polynomial map $g$ from the parameter space defining the model (a subset of $R^m$) to the parameter space over the observables (a subset of $R^n$).

We do not have satisfactory answers to these questions and so we can only give answers on small examples. We will focus on the zero-mean, three features, Naive Gaussian model presented in Section 3.3. This example shows how a cylinder algebraic decomposition can be used to answer these questions. More generally, finding a stratification for an arbitrary semialgebraic set, can be done in non-elementary complexity (e.g., Benedeti and Risler, 1990). This high complexity stems from applying $n$ steps of the doubly exponential ComputeCAD procedure (described in the appendix). A faster algorithm is not known to us.

Consider the example in Section 3.3. There are 4 cells of three dimensions denoted $C_{+--}$, $C_{-+-}$, $C_{--+}$, $C_{+++}$, 6 one dimensional cells which are the positive and negative parts of the three axis, denoted $C_{00+}$, $C_{00-}$, $C_{0+0}$, $C_{010}$, $C_{+00}$, $C_{-00}$, and one zero dimensional cell $C_{000}$—the origin. It is easy to see, in this example, that this cylinder algebraic decomposition is actually a stratification and that the singular points are the union of cells of dimensions 1 and 0. The cylinder algebraic decomposition for $f(x_1, x_2) = ax_2^2 + bx_2 + c - x_1$, on the other hand, discussed in the appendix, shows that lower dimensional cells need not be singular points, and that deciding whether cells are singular or not requires, in general, to examine their relation to higher dimensional cells—a task which seems computationally hard.

## 5 Concluding remarks

Hong's implementation of (partial) quantifier elimination QEPCAD for real closed fields (Version 19) allowed us to run the examples that we discussed in Section 3. However, when we slightly increased the complexity of the input, QEPCAD was running out of memory or we run out of patience. QEPCAD has many tuning parameters and we hope that they can provide extra power for slightly larger problems. Another possible source of improvements is the special form of our input formulae. For example, the Tarski sentences for model comparison between discrete Bayesian networks with hidden variables always produces multilinear polynomials (i.e., every variable appears with degree at most 1). Work on quantifier elimination of formulae of restricted form is a topic of research in symbolic algebra (e.g., Weisspfenning, 1988) .

## Acknowledgment

We are grateful to David Cox, Hoon Hong, and Bud Mishra for their helpful suggestions. We thank Hoon Hong for providing us with QEPCAD and useful examples. We thank David Heckerman, Judea Pearl, and Milan Studeny for discussions that motivated us to pursue this work, and to Thomas Richardson for providing us the references to work on the Heywood model.

# 6 Appendix

The presentation of Cylinder Algebraic Decomposition (*CAD*) and Quantifier Elimination (QE), which are the subjects of this section, is based primarily on Chapters 7 and 8 of Mishra (1993). Other excellent references are (Schwartz and Sharir, 1983) and a collection edited by Caviness and Johnson (1998) which reprints Tarski's and Collins' original papers (1951,1975) and provides a historical account of the development in real algebra and real geometry along with many important references which are omitted here. The entire theory is rooted in classic work on polynomials.

## 6.1 Basic facts about polynomials

The set of polynomials in variables $x_1, \ldots, x_n$ and coefficients in $S$ is denoted by $S[x_1, \ldots, x_n]$. Each polynomial in $S[x_1, \ldots, x_n]$ can be regarded as a polynomial in $x_n$ with coefficients in $S[x_1, \ldots, x_{n-1}]$. The notation $S[x_1, \ldots, x_{n-1}][x_n]$ expresses this idea. Throughout, to simplify our presentation, the set $S$ is assumed to be the set of polynomials of some maximal degree (possibly zero) with rational coefficients. When the maximal degree of the polynomials in $S$ is zero, we denote the coefficients with $Q$. The theorems that we cite, however, hold for more general sets of coefficients. For details, examine the aforementioned references.

The *Resultant* of of two polynomials $A(x)$ and $B(x)$ in $S[x]$ with degrees $m$ and $n$ respectively, denoted by $Resultant(A, B)$, is the determinant of the following $(m+n) \times (m+n)$ matrix over $S$ (called the *Sylvester* *matrix* of $A(x)$ and $B(x)$):

$$\begin{pmatrix} a_m & a_{m-1} & \cdots & a_0 & & & \\ & a_m & a_{m-1} & \cdots & a_0 & & \\ & & \cdots & \cdots & \cdots & & \\ & & & a_m & a_{m-1} & \cdots & a_0 \\ b_n & b_{n-1} & \cdots & b_0 & & & \\ & b_m & b_{n-1} & \cdots & b_0 & & \\ & & \cdots & \cdots & \cdots & & \\ & & & b_n & b_{n-1} & \cdots & b_0 \end{pmatrix}$$

which consists of $n$ staged rows of coefficients of $A$ and $m$ staged rows of coefficients of $B$.

**Theorem 3** *A polynomial $A(x) \in S[x]$ of degree at least two has a repeated factor iff Resultant$(A, A') = 0$ where $A'(x)$ is the (formal) derivative of $A(x)$.*

For example, if $A(x) = ax^2 + bx + c$ where $a > 0$, then $A'(x) = 2ax + b$, the Sylvester matrix is

$$\begin{pmatrix} a & b & c \\ 2a & b & \\ & 2a & b \end{pmatrix}$$

and $Resultant(A, A') = a(4ac - b^2)$. Consequently, $A(x)$ has a repeated factor iff its discriminant $b^2 - 4ac$ equals zero.

The next definition and theorem establish conditions for two polynomials $A(x)$ and $B(x)$ to have a common divisor of degree $i$.

The *i-th principal subresultant coefficient of A and B*, denoted by $PSC_i(A, B)$, is the determinant of the matrix $M_i^{(m+n-2i)}$ where $m$ is the degree of $A(x)$, $n$ is the degree of $B(x)$, $0 \leq i \leq min(m, n)$, and where the matrix $M_i^{(m+n-2i)}$ is obtained as follows: Start with the Sylvester matrix of $A$ and $B$. Remove the first $i$ rows that correspond to the coefficients of $A$ and the first $i$ rows that correspond to the coefficients of $B$. Remove the first $i$ columns and the last $i$ columns. (Note that the remaining matrix is a square matrix of size $m + n - 2i$). The following theorem, which demonstrates the importance of these definitions, can be found, for example, in Mishra (1993), Corollary 7.7.9.

**Theorem 4** *Let and $A(x)$ and $B(x)$ be univariate polynomials of positive degree m and n, respectively, with coefficients in S. Then, for all $0 < i \leq min(m, n)$, $A(x)$ and $B(x)$ have a common divisor of degree i iff*

$$(\forall j < i)\, [PSC_j(A, B) = 0] \wedge PCS_i(A, B) \neq 0.$$

For example, if $A(x) = ax^2+bx+c$ and $B(x) = 2ax+b$, then $M_1^{(m+n-2)}$ is a matrix of size $1 \times 1$ which equals the constant $2a$, and $PSC_1(A, B) = 2a$. Recalling, that $PSC_0(A, B) = a(4ac - b^2)$, we see that $A$ and $A'$ have a common divisor of degree 1 iff $b^2 - 4ac = 0$ and $a \neq 0$.

Let $A(x)$ be a polynomial of positive degree $n_1$ and $B(x)$ a polynomial of positive degree $n_2$. Define $n = n_1$ if $n_1 > n_2$ and $n = n_2$ otherwise. The *principle subresultant coefficient chain of A and B* is the sequence $PSC_{n+1} = 1, PSC_n = b_{n_2}$, where $b_{n_2}$ is the coefficient of $x^{n_2}$ in $B(x)$, $PSC_{n-1}(A, B), \ldots, PSC_i(A, B), \ldots, PSC_0(A, B)$. Theorem 4 tells us that by knowing this chain we can determine the degree of the common divisor of $A$ and $B$.

## 6.2 Cylinder Algebraic Decomposition

An element $u$ of $R$ is called a *real algebraic number* if for some nonzero polynomial $f(x) \in Q[x]$, $f(u) = 0$, where $Q[x]$ are the set of polynomials in $x$ with rational coefficients. Real algebraic numbers form a field. A real algebraic number $\alpha$ can be finitely represented as a triplet $(f(x), l, r)$ where $f$ is a polynomial in $Q[x]$ such that $f(\alpha) = 0$ and where $(l, r)$ is an interval that contains $\alpha$ and no other root of $f$. In this representation, addition, subtraction, multiplication, division, equality to zero, shrinking the interval $(l, r)$ by a factor of at least 2, normalizing $(l, r)$ such that $l$ and $r$ will have the same sign, are operations that can all be implemented in polynomial time (in the degree of $f$ and the log size of the coefficients of $f$). See Mishra (1993) for details. These operations on real algebraic numbers allow us to use finite representations of such (possibly irrational) numbers and explicate them in an arbitrary precision whenever needed.

**Definition:** A *decomposition* of $R^n$ is a finite collection of disjoint connected subsets of $R^n$ called *cells.* A *cylinder algebraic decomposition (CAD)* of $R^n$ is defined inductively as follows: When $n = 1$, *CAD* is a partition of $R$ into a finite set of real algebraic numbers, and into the finite and infinite open intervals bounded by these numbers. A *CAD* of $R^n$ is defined in terms of a *CAD* K' of $R^{n-1}$ via an auxiliary polynomial

$$g_{C'}(\overline{x}, x_n) = g_{C'}(x_1, \ldots, x_{n-1}, x_n) \in Q[x_1, \ldots, x_{n-1}][x_n]$$

one per $C' \in$ K'. The cells of K are of two kinds:

1. For each $C' \in$ K', we have $C' \times (R \cup \{\pm\infty\})$, namely, a *cylinder over C*.
2. For each $C' \in$ K', the polynomial $g_{C'}(p', x_n)$ has $m$ distinct real roots for each $p' \in C'$:

   $$r_1(p'), r_2(p'), \ldots, r_m(p'),$$

   each $r_i$ being a continuous function of $p'$. The following sectors and sections are cylinders over

**ALGORITHM ComputeCAD(F,j)**

**Input:** $F \subset Q[x_1, \ldots, x_j]$.
**Output:** $(K_j, \alpha_j)$ where $K_j$ is an F-sign-invariant *CAD* of $R^j$ and $\alpha_j$ is a set of algebraic sample points, one per cell in $K_j$.
**Recurse:** If $j > 1$, then do
$\Phi(F) := \Phi_1(F) \cup \Phi_2(F) \cup \Phi_3(F)$
$(K_{j-1}, \alpha_{j-1}) := ComputeCAD(\Phi(F), j-1)$,
else find the roots $r_1, \ldots, r_m$ of all polynomials in F and do
$K_1 := \{[-\infty, r_1), [r_1, r_1], (r_1, r_2), \ldots, (r_m, +\infty]\}$
$\alpha_1 := \{r_1 - 1, r_1, (r_1 + r_2)/2, \ldots, r_m, r_m + 1\}$
**Return** $(K_1, \alpha_1)$
**Lift:** For every cell $C_i \in K_{j-1}$ do
1. Compute the product of all polynomials in $F$ that do not vanish at the sample point $\alpha_i$ of $C_i$ and call the resulting polynomial $\pi(\alpha_i, x)$
2. Find the roots $r_1, \ldots, r_m$ of $\pi(\alpha_i, x)$
3. Set $K_{j,i} := \{\{C_i \times [-\infty, r_1)\}, \{C_i \times [r_1, r_1]\}, \{C_i \times (r_1, r_2)\}, \ldots, \{C_i \times (r_m, +\infty]\}\}$
**Comment:** $K_{j,i}$ are the cylinders over $C_i$.
4. Set $\alpha_{j,i} := \{(\alpha_i, r_1 - 1), (\alpha_i, r_1), (\alpha_i, (r_1 + r_2)/2), \ldots, (\alpha_i, r_m + 1)\}$
**Comment:** $\alpha_{j,i}$ are the algebraic sample points for the cylinders over $C_i$.

$K_j := \bigcup_i K_{j,i}$ ; $\alpha_j := \bigcup_i \alpha_{j,i}$
**Return** $(K_j, \alpha_j)$

Figure 2: An algorithm for computing a *CAD* of $R^j$

$C'$:

$$
\begin{aligned}
C_0^* &= \{(p', x_n) | x_n \in [-\infty, r_1(p')]\}, \\
C_1 &= \{(p', x_n) | x_n \in [r_1(p'), r_1(p')]\}, \\
C_1^* &= \{(p', x_n) | x_n \in [r_1(p'), r_2(p')]\}, \\
C_2 &= \{(p', x_n) | x_n \in [r_2(p'), r_2(p')]\}, \\
&\ldots \\
C_m &= \{(p', x_n) | x_n \in [r_m(p'), r_m(p')]\}, \\
C_m^* &= \{(p', x_n) | x_n \in [r_1(p'), +\infty\}
\end{aligned}
$$

□

An example of a *CAD* K of *R* via the two real roots $\alpha_1$ and $\alpha_2$ of $p(x) = ax^2 + bx + c$ (assuming they exist) is given by the cells $\{[-\infty, \alpha_1), [\alpha_1, \alpha_1]$, $(\alpha_1, \alpha_2), [\alpha_2, \alpha_2], (\alpha_2, \infty]\}$. This *CAD* has the property that the sign of $p(x)$ is constant in each cell of K. Consequently, determining whether $\exists x(p(x) \geq 0)$ is true reduces to the problem of checking one sample point from each cell. The sample points can be selected to be real algebraic points, e.g., $\alpha_1 - 1, \alpha_1, (\alpha_1 + \alpha_2)/2, \alpha_2, \alpha_2 + 1$. This example demonstrates how a *CAD* K can be used to determine whether a Tarski sentence is true. We now discuss how to construct such a sign invariant *CAD* and how to extend this construction to higher dimensions.

Let $F \subset Q[x]$ and K be a *CAD* of $R^n$. Then K is an *F-sign-invariant CAD* if for every $p(x) \in F$ and every $C \in K$ , either $p(x) > 0$ for all $x \in C$, or $p(x) = 0$ for all $x \in C$, or $p(x) < 0$ for all $x \in C$.

To find an F-sign-invariant *CAD* K of *R*, one constructs the auxiliary polynomial $\pi(x) = \Pi_{p(x) \in F} p(x)$, lists the real roots $\alpha_1, \ldots, \alpha_m$ of $\pi(x)$, and sets K to be these $m$ numbers joined with the corresponding $m + 1$ intervals. The key device for listing the real roots of an arbitrary polynomial in $Q[x]$ is an algorithm for ***root separation*** which constructs a set of intervals $(l_i, r_i)$, $i = 1, \ldots, m$, each containing exactly one root such that the representation of $\alpha_i$ is $(\pi(x), l_i, r_i)$. The algorithm to find such intervals is a binary search. Start with an interval $[-M, M]$ which contains all roots of $\pi(x)$. Divide the interval into two equal intervals ***left*** and ***right***. If ***left*** contains one root, stop dividing this interval, else continue recursively. Similarly, if ***right*** contains one root, stop dividing this interval, else continue recursively. The constant $M$ can be defined in terms of the coefficients of $\pi(x)$ and the test for the number of real roots in a given interval is based on Sturm's classical theory. For details consult Mishra (1993).

The extension to higher dimensions is achieved by a certain projection operator. For a set of polynomials $F = \{f_1, \ldots, f_s\}$ in $Q[x_1, \ldots, x_n]$, where $f_i^k$ is the coefficient of $x^k$ in $f_i$, the projection $\Phi(F)$ of $F$ on $\{x_1, \ldots, x_n\}$ is defined by $\Phi(F) = \Phi_1(F) \cup \Phi_2(F) \cup \Phi_3(F)$ where

$$
\begin{aligned}
\Phi_1(F) &= \{f_i^k(x_1, \ldots, x_{n-1}) | 1 \leq i \leq s, 0 \leq k \leq d_i\}, \\
\Phi_2(F) &= \{PSC_l^{x_n}(f_i(x_1, \ldots, x_n), D_{x_n}(f_i(x_1, \ldots, x_n))) \\
&\quad | 1 \leq i \leq s, 0 \leq l \leq d_i - 1\},
\end{aligned}
$$

$$\Phi_3(F) = \{PSC_l^{x_n}(f_i(x_1,\ldots,x_n), f_j(x_1,\ldots,x_n)) \mid 1 \le i < j \le s, 0 \le m \le \min(d_i,d_j)\}.$$

For example, for $f(x_1,x_2) = ax_2^2+bx_2+c-x_1$, we have $\Phi_1(\{f\}) = \{a,b,c-x_1\}$, $\Phi_2(\{f\} = \{4a^2((c-b^2/4a)-x_1)\}$ and $\Phi_3(F)$ is empty because F consists of only one polynomial. The importance of $\Phi(F)$ is demonstrated by the following theorem (Based on Chapter 8 in Mishra (1993)).

**Theorem 5 (Collins)** *Let $F \subset Q[x_1,\ldots,x_n]$ and $\Phi(F) \subset Q[x_1,\ldots,x_{n-1}]$ be the projection of F on $\{x_1,\ldots,x_{n-1}\}$. Let $K_{n-1}$ be a $\Phi(F)$-sign-invariant CAD of $R^{n-1}$ and $\alpha_i$ be an algebraic point from cell $C_i$ of $K_{n-1}$. Also, for every cell $C_i$ of $K_{n-1}$, let $\pi(\alpha_i,x)$ be the product of all polynomials in F that do not vanish at $\alpha_i$. Let $r_1,\ldots,r_m$ be the roots of $\pi(\alpha_i,x)$. Then, $K_n = \bigcup_i K_{n,i}$, where*

$$K_{n,i} := \{\{C_i\times[-\infty,r_1)\},\{C_i\times[r_1,r_1]\}, \{C_i\times(r_1,r_2)\},\ldots,\{C_i\times(r_m,+\infty]\}\},$$

*is a F-sign-invariant CAD of $R^n$ and $\{(\alpha_i,r_1-1),(\alpha_i,r_1),(\alpha_i,(r_1+r_2)/2),\ldots,(\alpha_i,r_m+1)\}$ are algebraic sample points for the cells of $K_n$.*

Theorem 5 prescribes a recursive algorithm to compute a F-sign-invariant *CAD* of a set F of polynomials in $Q[x_1,\ldots,x_n]$. We demonstrate this claim on a simple two dimensional example. The complete algorithm is given in Figure 2. Consider $F = \{f\}$, where $f(x_1,x_2) = ax_2^2+bx_2+c-x_1$. We have computed earlier and found that $\Phi(F) = \{a,b,c-x_1,4a^2((c-b^2/4a)-x_1)\}$. This set is in $Q[x_1]$. The roots of all polynomials in $\Phi(F)$ are $\{\alpha_1 = c-b^2/4a, \alpha_2 = c\}$. Note that, for the purpose of finding roots, we may as well replace $\Phi(F)$ with the (normalized) set $\{1, c-x_1, (c-b^2/4a)-x_1\}$ which has the same set of roots. In our example, these points are simply the $x_1$ coordinates of the minimum of the parabola (as a function of $x_2$) and its intersection with the axis $x_1$. Now, a $\Phi(F)$-sign-invariant *CAD* of $R$ consists of 5 cells, $\{([-\infty,\alpha_1),[\alpha_1,\alpha_1],(\alpha_1,\alpha_2),[\alpha_1,\alpha_2],(\alpha_2,+\infty]$, denoted by, $C_1,\ldots,C_5$, respectively. We can now apply Theorem 5 to construct the cells of a F-sign-invariant *CAD* of $R^2$. The only cell above $C_1$ is $C_1\times[-\infty,+\infty]$. To compute the cells above $C_5$ we compute the two roots $r_1$ and $r_2$ of $f(x_1,x_2)$ at $x_1=\alpha_2+1$ and the five cells are $\{C_5\times[-\infty,r_1),C_5\times[r_1,r_1],C_5\times(r_1,r_2),C_5\times[r_2,r_2],C_5\times(r_2,+\infty]\}$. All other cells over $C_2$,$C_3$ and $C_4$ are computed similarly. We denote with $C_{ij}$ the cells that are cylinders over $C_i$.

### 6.3 Quantifier Elimination

Collins' algorithm for constructing a cylinder algebraic decomposition provides a simple procedure for evaluating the truth value of a Tarski sentence $\psi$ and for generating a quantifier-free formula equivalent to $\psi$. In describing this procedure, we assume $\psi$ is given in prenex normal form, whereby all quantifiers appear in the beginning of the sentence, and only disjunctions, conjunctions, and negations are used. Every first order sentence can be rewritten in this way. In fact, in the case of Tarski sentences, we do not even need negation because $\neg(p(x)>0)$ can be replaced with $p(x)=0 \vee p(x)<0$.

Suppose $\psi$ is a Tarski sentence given in the form $(Q_1x_1)\cdots(Q_nx_n)M(x_1,\ldots,x_n)$ where $Q_i$ is either a universal quantifier $\forall$ or an existential quantifier $\exists$, and where $M$ is a quantifier-free formula involving a set of polynomial $F \subset Q[x_1,\ldots,x_n]$. Collins' algorithm for evaluating the truth value of $\psi$ is as follows. Construct an F-sign-invariant *CAD* K of $R^n$ in the order $x_1,\ldots,x_n$. The resulting CAD forms a tree. The leaves of the tree represent the cells of $K$. The nodes in level $i$ represent a CAD of $R^i$. The children of a node representing a cell $C$ in a CAD of $R^i$ represent the cylinders over $C$ in the CAD of $R^{i+1}$. The tree is regarded as an and-or tree where in level $i$ there is an and if $Q_i=\forall$ and an or if $Q_i=\exists$.

Evaluating the truth value of $\psi$ is done recursively as follows. First, each leaf, representing cell $C$, is evaluated to true or false depending on whether $M(\alpha)$ is true or false, where $\alpha$ is the sample point of $C$. The truth value of a node $C$ in level $i$ is determined by the truth value of its children. If $Q_i=\forall$, then $C$ evaluates to true iff all its children evaluate to true, and if $Q_i=\exists$, then $C$ evaluates to true iff at least one of its children evaluates to true. The sentence $\psi$ is true iff the root of the and-or tree evaluates to true. Furthermore, each leaf represents a cell in a CAD of $R^n$, which is described with a quantifier-free Tarski sentence. The sentence $\psi$ is equivalent to the quantifier-free formula obtained by applying the and/or connectives, as prescribed by the and-or tree, to the quantifier-free formula that describe the leaves.

In the example of Section 6.2, the nodes in the first level correspond to the cells $C_1,\ldots C_5$, and the nodes in the second level correspond to the cells $\{C_{i,j}\}$. Node $C_1$ has one child $C_{1,1}$, node $C_2$ has three children $C_{2,1},C_{2,2},C_{2,3}$, and nodes $C_2$, $C_3$, and $C_5$, each have five children. The formula $(\exists x_1)(\forall x_2)ax_2^2+bx_2+c-x_1>0$ is true because $C_{1,1}$ evaluates to true and so $C_1$ evaluates to true as well. Consequently, since $x_1$ is existentially quantified, the root is evaluated to true. The cell $C_1$ is given by the quantifier-free formula $x_1 < c-b^2/4a$ which has a clear geometrical interpretation; The formula is true whenever $x_1$ is smaller than the minimum of the parabola $ax_2^2+bx_2+c$.